\pdfoutput=1

\documentclass[11pt]{article}

\usepackage[preprint]{acl}

\usepackage{times}
\usepackage{amsmath}
\usepackage{xcolor}
\usepackage{soul}

\sethlcolor{green!65} 
\usepackage{latexsym}
\usepackage{colortbl}
\usepackage{color}
\definecolor{LLGray}{gray}{0.91}
\definecolor{Greenish}{rgb}{0.10,0.60,0.30}
\usepackage{enumitem}

\usepackage[T1]{fontenc}

\usepackage[utf8]{inputenc}

\usepackage{microtype}

\usepackage{inconsolata}

\usepackage{graphicx}

\usepackage{amssymb}

\title{A Multimodal Recaptioning Framework to Account for Perceptual Diversity Across Languages in Vision-Language Modeling}

\author{Kyle Buettner\textsuperscript{\rm 1},      Jacob T. Emmerson\textsuperscript{\rm 2},
    Adriana Kovashka\textsuperscript{\rm 1,2}\\
\textsuperscript{1}Intelligent Systems Program, \textsuperscript{2}Department of Computer Science,
University of Pittsburgh \\
{\tt buettnerk@pitt.edu, jte27@pitt.edu, kovashka@cs.pitt.edu}\\
\small \href{https://krbuettner.github.io/PerceptualUnderstandingAcrossLanguages}{https://krbuettner.github.io/PerceptualDiversityAcrossLanguages}
}

\begin{document}
\maketitle

\begin{abstract}

    When captioning an image, people describe objects in diverse ways, such as by using different terms and/or including details that are perceptually noteworthy to them. Descriptions can be especially unique across languages and cultures. Modern vision-language models (VLMs) gain understanding of images with text in different languages often through training on machine translations of English captions. However, this process relies on input content written from the perception of English speakers, leading to a perceptual bias. In this work, we outline a framework to address this bias. We specifically use a small amount of native speaker data, nearest-neighbor example guidance, and multimodal LLM reasoning to augment captions to better reflect descriptions in a target language. When adding the resulting rewrites to multilingual CLIP finetuning, we improve on German and Japanese text-image retrieval case studies (up to +3.5 mean recall, +4.4 on native vs. translation errors). We also propose a mechanism to build understanding of object description variation across languages, and offer insights into cross-dataset and cross-language generalization.

\end{abstract}

\section{Introduction}

    \begin{figure}[t]
    \centering
    \includegraphics[width=0.96\linewidth]{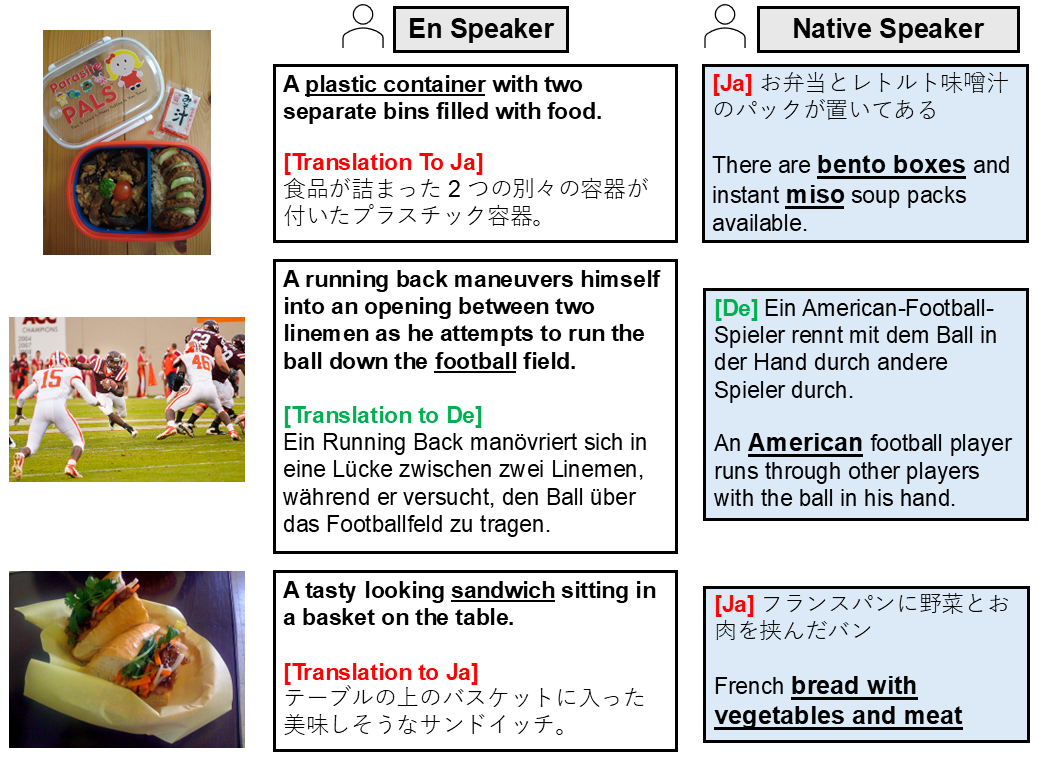}
    \vspace{-3mm}
    \caption{\textbf{English captions (and their translations) do not capture the perceptual diversity of object and scene descriptions in other languages.} They often fail to include cultural terms (\textit{e.g.} \textit{bento box}) and miss differences in native perspective (\textit{e.g.} German emphasis of \textit{\underline{American} football}). More subtlely, we find that they differ from cross-language captions in the use of common nouns, for instance in Japanese STAIR \citep{yoshikawa2017stair} where \textit{bread} is more frequently described, especially with its contents (\textit{e.g.} \textit{vegetables}). Our multimodal recaptioning method considers these differences to enhance cross-lingual training data generation.
    }
    \label{intro_fig}
\end{figure}


    People vary in how they describe the same visual scene. They may note different foreground or background objects (\emph{e.g.} \textit{person} vs. \textit{sky}). Objects may be described apart or grouped under umbrella terms (\emph{e.g.} \textit{sofa}, \textit{table}, and \textit{chair} vs. \textit{furniture}). The same object may be noted with a base term (\emph{e.g.} \textit{dog}), hypernyms (\emph{e.g.} \textit{animal}), hyponyms (\emph{e.g.} \textit{Boston Terrier}), or synonyms (\emph{e.g.} \textit{canine}). Context, like attributes (\emph{e.g.} \textit{yellow}), may be described if noteworthy or unusual, and captions may vary in detail. 
    
    Differences are especially unique \textit{across languages} \citep{nguyen2024multilingual, Ye_2025_CVPR}, where speakers have diverse perspectives, knowledge, and experiences that contribute to language production. For instance, as shown in Fig. \ref{intro_fig}, an English speaker may describe an image as containing a \textit{plastic container}, while a Japanese speaker may perceive a \textit{bento box}. The diversity is sometimes subtle, and reflected in object descriptions having different frequencies, abstraction, and usage patterns. For example, we find Japanese captions to mention \textit{sunglasses} 5.6$\times$ more often than English ones, potentially due to their relative uncommonness in Japan (thus noteworthiness). There is notably a distribution gap between English captions and captions from other languages in their \textit{perceptual diversity}. 
    
    With the rise of multilingual vision-language models (VLMs) \citep{zhai2023sigmoid, carlsson2022cross, yue2024pangea, chen2022pali, Chen_2024_CVPR, geigle2023mblip}, machine translation from English has been used to generate cross-lingual data. A key observation is that \textit{machine translation does not significantly adapt semantic content}. It relies on source object naming and context, leading to an English perceptual bias that limits understanding of native text in other languages. To reduce bias, text can be diversified with strategies like paraphrasing and general captioning. We explore such options, but find they achieve limited understanding of culture-specific perceptual differences.

    As a more compelling strategy to reduce perceptual bias, we propose a multimodal recaptioning framework that alters object descriptions to reflect properties in a target language. We simply incorporate a small amount of reference data and image context in the prompting of a multimodal LLM to infer how concepts are described cross-language, and then produce rewrites. This process is guided by reference captions from similar scenes, selected as nearest neighbors in image similarity space.     
    
    Rewrites are then integrated as random augmentations in the training of mCLIP \cite{chen2023mclip}, a multilingual image-text retrieval model. We compare the use of rewrites generated through this \textit{targeted image recaptioning} to the use of captions produced from non-targeted prompts that encourage diversity (i.e. paraphrasing and general recaptioning). Performance is evaluated with Japanese STAIR \citep{yoshikawa2017stair}, German Multi30k \citep{elliott2016multi30k}, and XM3600 \citep{thapliyal2022crossmodal}, including in cross-dataset and cross-language settings. Notably, targeted image recaptioning improves default training by up to +2.4 mean recall and outperforms  both diverse paraphrasing and image captioning on native vs. translation error sets which isolate perceptual differences between languages by up to +4.4. In combination, all rewrites improve mCLIP by +3.5. We share further insights by highlighting key differences in object term distributions across datasets. 
        
    In summary, along with our framework, our main contributions are insights into these questions: 
    \begin{itemize}[noitemsep,nolistsep]
        \item In which contexts is targeted image recaptioning beneficial?
        \item How do differences in object descriptions uniquely manifest across languages?
        \item Does understanding of perceptual diversity gained from one target language dataset generalize to other datasets and languages?
    \end{itemize}

    
\section{Related Work}

    \noindent \textbf{Perceptual differences in object descriptions.} Much progress in computer vision has been driven by models trained on datasets with mostly English text, such as the CLIP dataset \citep{radford2021learning}, or on datasets with English noun hierarchies, such as ImageNet \citep{deng2009imagenet}.  
    The underlying concept representations are thus biased towards the details English speakers find salient, and the entry-level categories English speakers use to name objects \citep{ordonez2013large}. With English conceptual understanding lacking universality \cite{liu2021visually}, models may fail to adequately capture perceptual diversity across cultures (\emph{e.g.} representation of the Japanese \textit{koto} instrument). There is much research on cultural differences in perception, such as on attention to foreground vs. background \cite{nisbett2013culture} and on attributes prescribed to objects because of grammar \cite{boroditsky2006linguistic}. Recent work has found that perceptual diversity in multilingual datasets helps English vision tasks \cite{nguyen2024multilingual}. Our work addresses English perceptual bias to achieve greater understanding of native speaker text in other languages.

    \noindent \textbf{Multilingual, vision-language modeling.} Recent works have moved to instill popular English-centric VLMs and multimodal LLMs with multilingual capabilities \citep{zhai2023sigmoid, carlsson2022cross, yue2024pangea, chen2022pali, Chen_2024_CVPR, geigle2023mblip}. In the absence of native speaker data, it is common to machine-translate captions \citep{carlsson2022cross} and instructions \citep{yue2024pangea}, though the resulting text carries an English bias. We show that \textit{targeted, image recaptioning} can address this bias, and improve multilingual CLIP retrieval \citep{chen2023mclip} on two languages (German/Japanese). Our work fits with recent VLM works that effectively leverage LLMs to generate diversity in text for retrieval/classification \citep{fan2023improving} and compositional understanding \citep{doveh2023teaching, doveh2023dense}. It also relates to aligning multiple texts to an image \citep{sarto2023positive, bulat2024fff}, but past work does not consider differences across languages. We aim to align multiple, perceptually diverse views to an image through random sampling of text in training.

    
    \noindent \textbf{Machine translation and image captioning.} Gaps have been shown in image-text retrieval when using translated vs. native speaker training data \cite{kadar2018lessons}. \citet{buettner2024quantifying} show that \textit{text-only paraphrasing} techniques can partially address gaps (+1.3 performance). Our perspective is that text changes need to be \textit{visually} driven as speakers across cultures may uniquely focus on different parts of the image, which paraphrasing does not address. We thus employ new \textit{image}-based reference sampling and targeted, \textit{image} recaptioning techniques, and further consider a greater scope of languages and datasets. Other work \cite{yang2023re} uses $k$-NN in image embedding space to help retrieve knowledge for few-shot English captioning. Our method is unique as the nearest image neighbor guides creation of \textit{perceptually diverse}, \textit{cross-lingual} data to be used as training augmentations in \textit{image-text retrieval}. \citet{ramos-etal-2024-paella} train a model for multilingual captioning using retrieval with reference translations. We alternatively show importance of a reference set with captions directly produced by native speakers of another language, and our recaptioning does not require training (just prompting). Our work is loosely related to nearest-neighbor machine translation \citep{khandelwal2020nearest} and multimodal machine translation \citep{yao2020multimodal}. Unlike in machine translation, we significantly adjust caption content to address perceptual bias.
     

\section{Experimental Methodology}

    \begin{figure*}[t]
    \centering
    \includegraphics[width=0.96\linewidth]{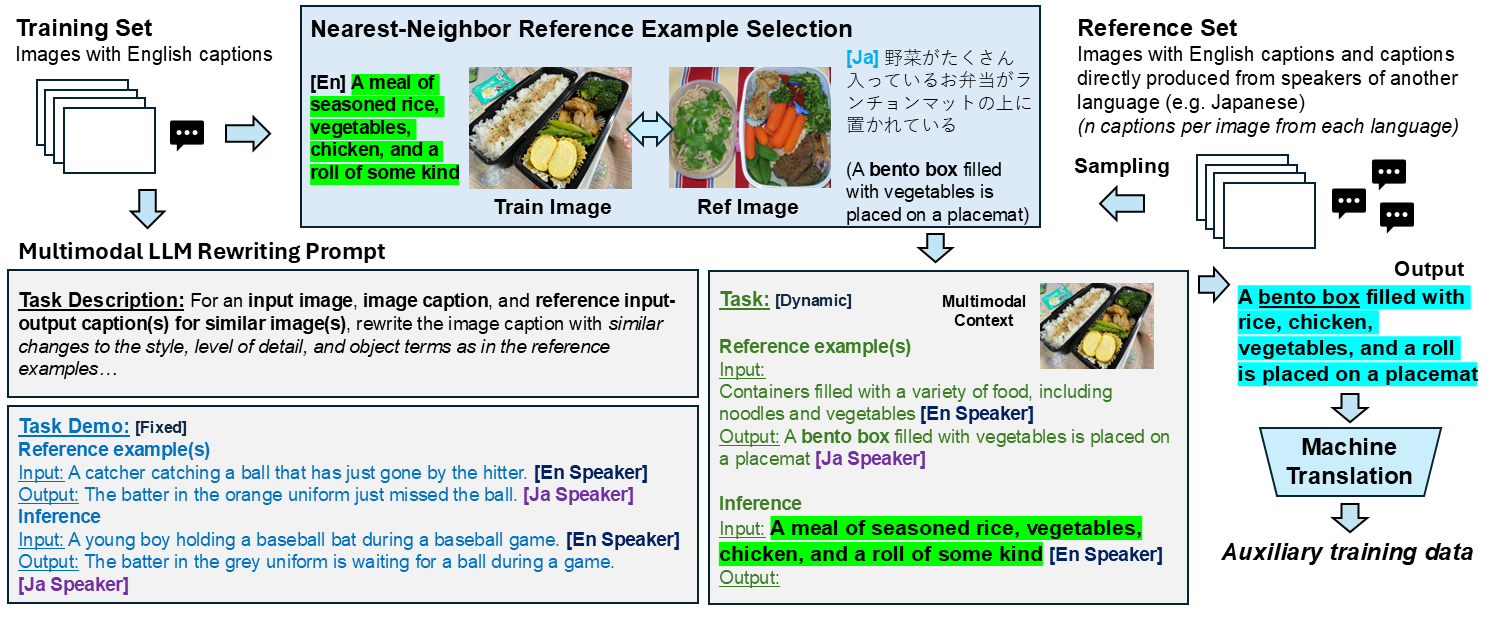}
    \vspace{-3mm}
    \caption{\textbf{Our multimodal, LLM-based recaptioning method to adapt object descriptions before translation.} For a set of images with only English captions, we generate new captions which better represent perceptual properties of a target language (\emph{e.g.} Japanese). Each generation is guided by a reference example selected as the nearest neighbor in image similarity from a small set of native speaker data. Using the prompt shown, the multimodal LLM leverages the reference example and image context to infer targeted changes. This example shows the model adding the cultural term \textit{bento} while also listing foods relevant to the input image. Text in brackets is not in the prompt.}
    \label{prompt_fig}
\end{figure*}
    
    We consider an image-text retrieval case study where the text queried and retrieved come from native speakers of a target language (\emph{e.g.} Japanese, German). The goal is to design a multimodal framework that improves VLM understanding of perceptually diverse text across languages. Specifically, we aim to enhance training with machine translation of English text, which is a practical strategy when a \emph{large} amount of target language text from native speakers is unavailable. The challenge 
    is that caption properties, such as which objects are mentioned, the level of detail or context described, and use of certain synonyms, hypernyms, or hyponyms, are biased towards English perception.

    To address this bias, for a given input caption and image, we propose to use a multimodal LLM (\textit{Llama-3.2-11B-Vision-Instruct}) to produce rewrite(s), specifically by leveraging multimodal context and reference examples selected by image similarity. Outlined in this section are our framework's retrieval model (Sec. \ref{prelim}),  mechanisms for generating data with adequate diversity (Sec. \ref{recap}), and strategy to identify object description differences between datasets (Sec. \ref{recipe}).

    \subsection{Preliminaries: Multilingual CLIP}
        \label{prelim}

        We select the retrieval model to be multilingual CLIP or \emph{mCLIP} \citep{chen2023mclip}, due to its support for multiple languages and CLIP's success as a VLM. This method extends the mostly English CLIP by replacing the text encoder with a multilingual model, XLM-R \citep{conneau-etal-2020-unsupervised}, and training lightweight projection layers to align multilingual text embeddings to CLIP image and text embeddings. We are primarily interested in \textit{finetuning} to adapt mCLIP's alignment of images and captions. We specifically train the image-to-text (I2T) and text-to-image (T2I) matching losses in \citet{chen2023mclip}, which operate over a batch of size $N$ where each sample $k$ has an image $i_k$ and text $t_k$. Shown in Eqs. \ref{i2t} and \ref{t2i} are these losses with the multilingual text encoder $f$, the CLIP image encoder $g$, a temperature $\tau$, and a similarity function $\langle \rangle$ (cosine): 

        \begin{equation}
            \label{i2t}
            \mathcal{L}_{I2T} = - \frac{1}{N} \sum_{k=1}^{N} \log \frac{\exp ({\langle f(t_k), g(i_k)\rangle/\tau})}{\sum\limits_{n=1}^{N} \exp ( \langle f(t_n), g(i_k) \rangle/\tau )}
        \end{equation}

        \begin{equation}
            \label{t2i}
            \mathcal{L}_{T2I} = - \frac{1}{N} \sum_{k=1}^{N} \log \frac{\exp ({\langle f(t_k), g(i_k)\rangle/\tau})}{\sum\limits_{n=1}^{N} \exp ( \langle f(t_k), g(i_n)\rangle/\tau)}
        \end{equation}

        The overall loss is $\mathcal{L}$ = $\frac{1}{2} $($\mathcal{L}_{I2T}$ + $\mathcal{L}_{T2I}$).

    \subsection{Multimodal Recaptioning}
        \label{recap}

        \noindent \textbf{Main approach.} Assume we have a dataset $\mathcal{D}_{train}$ of image-caption pairs with text in a source language $\mathcal{L}_{src}$ (i.e. English). We also have a target language $\mathcal{L}_{tgt}$ for which we wish to generate training data (\emph{e.g.} Japanese/German). We propose to alter the object descriptions of captions in $\mathcal{D}_{train}$ before machine translation to $\mathcal{L}_{tgt}$ for enhanced diversity. A natural approach is to paraphrase English text to produce alternative descriptions of objects (\emph{e.g.} calling a \textit{Boston Terrier} a \textit{dog}). However, it is difficult to infer from only the source text adequate culture-specific terms (e.g. \textit{yakitori} vs. \textit{chicken}). In addition, what may be deemed salient by an English speaker may exclude salient objects often written in text from a speaker of another language.  

        We reason that image context, along with effective guidance from native speaker data, are needed to adequately diversify captions. 
        We thus propose to use a multimodal LLM, with strong reasoning capabilities, to rewrite each training caption in a targeted manner. The idea is to leverage a small reference set $\mathcal{D}_{ref}$, disjoint from $\mathcal{D}_{train}$, which consists of images and captions produced in English \textit{and} from native speakers of $\mathcal{L}_{tgt}$ (\emph{e.g.} German/Japanese). A single input-output \textit{guidance example} is selected from $\mathcal{D}_{ref}$ to instruct the LLM on how an (input) English caption can be rewritten as an (output) caption from a native speaker of $\mathcal{L}_{tgt}$. The multimodal LLM then reasons how reference descriptions can apply to a new image. 
        
        We specifically use LLaMA 3.2 \citep{touvron2023llama}, with the instruction in Fig. \ref{prompt_fig}. Provided is a task description, demo for formatting (constant across inferences), the training image \textit{and} caption, and the sampled reference texts for the given input. Note in the example how the reference output shows description of \textit{bento box}, and that the LLM generalizes its use in the rewrite while also including input image-relevant ingredients (\emph{e.g.} \textit{rice}, \textit{chicken}). 
        The guidance example is translated to English (with Google Translate), and the LLM produces output in English. We recaption in English, then machine-translate, to ensure quality in a language with which we have familiarity.  
        
\begin{figure}[t]
    \centering
    \includegraphics[width=0.94\linewidth]{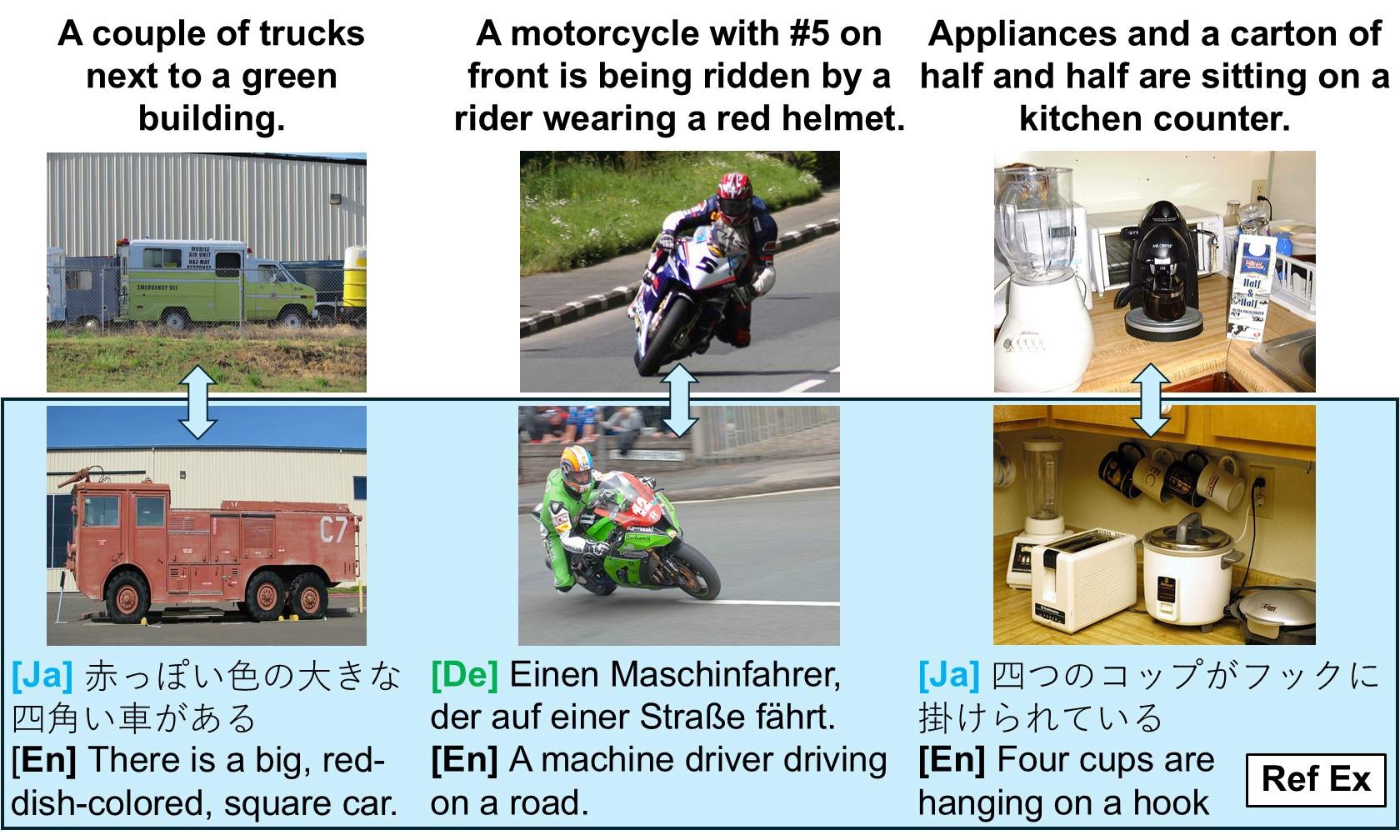}
    \vspace{-2mm}
    \caption{\textbf{Guidance from nearest neighbors can reveal subtle differences in object naming.} A reference is chosen based on image similarity to acquire diverse descriptions of related concepts and scenes. 
    Notice how \textit{truck} may be described loosely as \textit{car} in Japanese. Similarly, there are differences in object grouping and objects that are deemed salient (described).
    }
    \label{nn_fig}
\end{figure}

        \noindent \textbf{Choosing guidance examples.} Reference examples are selected with the idea that images with similar content may be described similarly. Thus if there is native speaker data for a similar image, object description properties can be inferred for the image to be recaptioned. We choose the nearest neighbor in image feature space as the guidance example, using image embeddings obtained from zero-shot mCLIP. This process captures culture-specific terms like in Fig. \ref{prompt_fig} and also more subtle differences across languages, as shown in Fig. \ref{nn_fig}. 

        \noindent\textbf{Other approaches.} We also consider non-targeted methods (without use of a reference set), given that there is general variability of object descriptions across languages. We propose a multimodal captioning strategy where given an image and caption, we encourage the LLM to rewrite the caption with differences in phrases, sentence structure, semantic content, which objects are described, and/or level of detail. Otherwise the format is the same as the targeted scenario, thus ablating the impact of reference guidance. Then as a baseline similar to prior work \citep{fan2023improving, buettner2024quantifying}, we prompt LLaMA to produce a paraphrase that reflects diversity in how speakers around the world describe objects across languages. We refer to the above two strategies as \textit{Diverse Image Recaptioning} and \textit{Diverse Paraphrasing}, respectively, and compare to our proposed \textit{Targeted Image Recaptioning}, when used separately and in conjunction. The prompts are in Appendix \ref{app_prompt}.

        \noindent \textbf{Incorporating rewrites into training.} To use rewrites (after machine translation), we leverage a similar mechanism to \citet{fan2023improving}, where generated captions are treated as random augmentations in retrieval training. This simple mechanism encourages images to match to multiple, perceptually diverse views over the course of training. Referring to Eqs. \ref{i2t} and \ref{t2i}, instead of using the original batch $B$ of image-text pairs ($i_k$, $t_k$), a new batch $B'$ is constructed consisting of image-text pairs ($i_k$, $t'_k$). Each $t'_k$ is sampled from a uniform distribution of all possible positives for an image (i.e. the original caption $t_k$ and each possible rewrite $p_k^i$). Formally, this process is shown in Eq. \ref{aug} for $n$ positive rewrites: 
        \begin{equation}
            \label{aug}
            t'_k \sim \text{Uniform}([t_k, p_k^1, ..., p_k^n])
        \end{equation}

        When one rewriting strategy is used, $n$=1, meaning that approximately 50\% of training instances are rewrites. We explore up to $n$=3 when considering all 3 rewrite strategies in combination. 
          
    \subsection{Identifying Object Description Differences Across Languages}
        \label{recipe}

         It remains an open question how object description differences manifest across inter-language datasets. Specific questions include how often certain synonyms or hyponyms are used (\emph{e.g.} in Fig. \ref{nn_fig}, \textit{truck} vs. \textit{car} for \textit{vehicle}), whether grouping terms are used (\emph{e.g.} \textit{furniture}), and if certain terms are
         used significantly more in one language than another. To explore, we design a strategy with WordNet \citep{miller1995wordnet} to create object description distributions.
        
         Consider a set $\mathcal{H}$ consisting of ``supercategories'' which cover various objects. In this study, we choose $\mathcal{H}$ to be $\{$person, conveyance, furniture, animal, container, food, device$\}$, representing common objects in COCO. Then given source language captions $\mathcal{D}_{src}$ and target language captions that have been translated to English $\mathcal{D}_{tgt}$, we process each caption in $\mathcal{D}_{src}$ and $\mathcal{D}_{tgt}$ with a SpaCy part of speech tagger (\textit{en\_core\_web\_sm}, v3.6.1) to identify nouns. Each noun is mapped to the probable WordNet synset (the first noun synset definition listed). Then if possible, we match each mapped synset to  the closest synset of a top-level term in $\mathcal{H}$ by performing a closure of hypernyms. We compare term frequencies under each supercategory in Sec. \ref{results} to learn about differences across languages.

\section{Experimental Settings}

 \setlength{\tabcolsep}{6pt}
\begin{table*}
    \begin{center}
    \renewcommand{\arraystretch}{0.75}
    \begin{tabular}{c||ccc|ccc|c}
    \hline
        \textbf{Method} & \multicolumn{3}{c|}{\textbf{I2T Retrieval}} & \multicolumn{3}{c|}{\textbf{T2I Retrieval}} &  \textbf{Mean} \\
      & \small R@1 & \small R@5 & \small R@10  & \small R@1 & \small R@5 & \small R@10 & \textbf{Recall}  \\
   
  \hline
 \multicolumn{8}{c}{\small \textbf{Train:} English COCO (to Japanese) / \textbf{Eval}: STAIR (Japanese)}\\
\hline
 \rowcolor{LLGray} \small mCLIP &  \small 10.2 & \small 25.0  & \small 33.9& \small 9.2 & \small 23.2 & \small 31.9  & \small 23.0   \\
    
\rowcolor{LLGray} \small FT on Japanese Data MT from English  &  \small 20.2 & \small 43.0  & \small 54.7 & \small 19.4 & \small 41.9 & \small 53.3 & \small 39.3   \\
    
     \small  + Rewrites from Diverse Paraphrasing &  \small 21.7 & \small 45.1  & \small 56.7  &  \small 20.6 & \small 43.9 & \small 55.4 & \small 40.6   \\
     \small + Rewrites from Diverse Image Recaptioning  &  \small 22.0 & \small 45.3  & \small 57.1 & \small 20.7 & \small 44.1 & \small 55.6  & \small 40.8  \\  
     \small \textbf{+ Rewrites from Targeted Image Recaptioning} &  \small \textbf{22.5} & \small \textbf{46.5}  & \small \textbf{58.1}  & \small \textbf{21.4} & \small \textbf{45.0} & \small \textbf{56.7}  & \small \textbf{41.7}   \\

     \rowcolor{LLGray} \small FT on Japanese Data from Native Speakers &  \small 24.8 & \small 50.2  & \small 62.0 & \small 24.3 & \small 49.4 & \small 61.2  & \small 45.7     \\  

   \hline
    \multicolumn{8}{c}{\small \textbf{Train:} English Flickr30k (to German) / \textbf{Eval}: Multi30k (German)}\\
    \hline
     \rowcolor{LLGray}  \small mCLIP &  \small 13.8 & \small 31.6  & \small 41.8  & \small 13.0 & \small 30.6 & \small 40.7 & \small 28.6 \\

    \rowcolor{LLGray}  \small FT on German Data MT from English &  \small 22.3 & \small 45.4  & \small 56.4& \small 21.5 & \small 44.4 & \small 55.6   & \small 40.9  \\
    
    \small  + Rewrites from Diverse Paraphrasing &  \small 22.5 & \small 46.0  & \small 57.1  & \small 21.7 & \small 44.9 & \small 56.2 & \small 41.4  \\
    
    \small + Rewrites from Diverse Image Recaptioning &  \small \textbf{22.7} & \small 46.3  & \small 57.5 & \small 22.1 & \small 45.5 & \small 56.8 & \small 41.8  \\
    
    \small \textbf{+ Rewrites from Targeted Image Recaptioning} &  \small \textbf{22.7} & \small \textbf{46.5}  & \small \textbf{57.8} & \small \textbf{22.3} & \small \textbf{45.8} & \small \textbf{57.1}   & \small \textbf{42.1}   \\

    \rowcolor{LLGray} \small FT on German Data from Native Speakers &  \small 22.5 & \small 46.1  & \small 57.6  & \small 22.1 & \small 45.9 & \small 57.1 & \small 41.9  \\   

   \hline 
   
    \end{tabular}
    \end{center}
    \vspace{-0.3cm}
    \caption{\textbf{Targeted image recaptioning is the most beneficial augmentation in text-image retrieval on native speaker data from STAIR (Japanese) and Multi30k (German).} The $+$ symbol indicates that data is added as augmentations to the ``FT on X Data MT from English'' setting. FT=finetuned, MT=machine-translated. }
   \label{table:main}
\end{table*}

    \noindent \textbf{Datasets.} We train on English COCO and evaluate on Japanese STAIR \cite{yoshikawa2017stair}, which includes Japanese captions from native speakers for COCO images. Similarly, we train on English Flickr30k and evaluate on German Multi30k \cite{elliott2016multi30k}, which includes German captions from native speakers for Flickr images. The English captions serve as input to recaptioning and machine translation for retrieval training. Disjoint sets of native captions are used in targeted recaptioning. Specifically, in both case studies, there are 5 English caption sets (5 captions per image). STAIR contains 5 Japanese captions for each COCO image, and Multi30k contains 5 German captions for each Flickr image. We randomly split the 5 sets for each dataset/language into reference, training, and evaluation sets, so there is disjoint data for the targeted method. The image split sizes for STAIR are 9,666/73,117/10,668 and for Multi30k are 9,666/9,666/10,668. For each image, our targeted method samples 1 caption from the 5 English reference sets and 1 caption from the 5 cross-language sets. Evaluation is averaged across the 5 test sets. For cross-dataset experiments, we evaluate on XM3600 \citep{thapliyal2022crossmodal}, as it includes native speaker data for 36 languages. We do not train on XM3600 due to size (3,600 images only). We report results for languages related to the ones for which we have native speaker data.

    \noindent \textbf{Recaptioning.} Rewrite generation is performed with the multimodal LLaMA 3.2 (\textit{Llama-3.2-11B-Vision-Instruct}). A temperature of 0, seed of 42, and max tokens 448 are used for all experiments.

    \noindent \textbf{Translation.} To generate cross-lingual training data, we use two models from HelsinkiNLP \citep{tiedemann-thottingal-2020-opus}, \textit{opus-tateoba-en-ja} for Japanese and \textit{opus-mt-en-de} for German, as these are amongst the most downloaded on HuggingFace. We also test No Language Left Behind \citep{costa2022no} in Appendix \ref{nllb}.  Translations of English captions, including rewrites, are generated with greedy decoding at a max token count of 200. We notably choose a much higher token count in translation than the paraphrasing work of \citet{buettner2024quantifying}, as it is found to significantly improve the quality of translation. 

    \noindent \textbf{Retrieval.} For \textit{training}, results are collected with the settings in \citet{chen2023mclip} (batch size 512, learning rate 0.001, 30 epochs, LAMB optimizer, temp. 0.07) on 1 NVIDIA A100 GPU. For \textit{evaluation}, I2T and T2I retrieval scores are calculated as recall@1/5/10. The mean of these six scores, termed mean recall \citep{chen2023mclip}, is calculated and reported as averages over the 5 test sets. 

    \noindent \textbf{Native vs. translation error sets.} We construct other test sets that isolate differences from using translation (with English perceptual bias) vs. native speaker data. 
    We train mCLIP models with captions translated from English to Japanese/German and with native Japanese/German captions. Then we collect I2T/T2I cases that the native models correctly retrieve within 10 samples, but the translation models get incorrect@10, as we reason these cases include errors that would be addressed with understanding of perceptual diversity. We refer to these collectively as \textit{Native vs. Translation Error Sets}. Evaluation considers retrieval over the full test sets, but mean recall is only calculated across cases in these sets. The I2T/T2I sample counts are 1,409/1,391 for STAIR and 994/946 for Multi30k. This evaluation is shown in Table \ref{table:nds}, while all other tables follow overall retrieval evaluation.

\section{Results and Analysis}

    \label{results}

         We evaluate \textit{Targeted Image Recaptioning} vs. \textit{Diverse Image Recaptioning} vs. \textit{Diverse Paraphrasing}; the latter represents baselines \citep{fan2023improving, buettner2024quantifying}. We baseline mCLIP without finetuning, and mCLIP finetuned on data that has been machine-translated from English to the target language (without recaptioning). The datasets from rewrite strategies are incorporated into finetuning as augmentations that are randomly sampled with original machine translations. As a reference, we evaluate finetuning with captions from speakers of each language (Native Ja/De), though this is not a strict upper bound since more data is used in rewrite settings. We also test combinations of all rewrite strategies. 

         \subsection{In which contexts is targeted image recaptioning beneficial?} 
                           
 \setlength{\tabcolsep}{0.6pt}
\begin{table}[t]
    \begin{center}
    \renewcommand{\arraystretch}{0.82}
    \begin{tabular}{c|ccc|ccc|c}
    \hline
      \textbf{Method}  & \multicolumn{3}{c|}{\textbf{I2T Retrieval}} & \multicolumn{3}{c|}{\textbf{T2I Retrieval}} & \textbf{Mean} \\
      & \small R@1 & \small R@5 & \small R@10  & \small R@1 & \small R@5 & \small R@10 & \textbf{Recall}  \\
    
    \hline
    \multicolumn{8}{c}{\small \textbf{Train:} English COCO / \textbf{Eval}: STAIR (Japanese)}\\
    \hline
    
    \rowcolor{LLGray} \small FT (MT Data)  & \small 0.0 & \small 0.0 & \small 0.0 & \small 0.0  & \small 0.0 & \small 0.0 & \small 0.0   \\
    
     \small  +Paraphrase  & \small 0.5 & \small 8.9 & \small 26.6  & \small 0.4 & \small 7.5 & \small 24.2 & \small 11.4    \\
     \small +Diverse Recap  & \small 0.7 & \small 8.8 & \small 26.4 & \small 0.9 & \small 9.5 & \small 25.0 & \small 11.9   \\
     \small +Tgt Recap   & \small \textbf{1.3} & \small \textbf{14.1} & \small \textbf{32.4} & \small \textbf{1.4} & \small \textbf{13.4} & \small \textbf{35.2} & \small \textbf{16.3} \\
     \rowcolor{LLGray} \small FT (Native Ja) & \small 16.0 & \small 60.0 & \small 100.0  & \small 16.4 & \small 61.1 & \small 100.0 & \small 58.9  \\

    \hline
    \multicolumn{8}{c}{\small \textbf{Train:} English Flickr30k / \textbf{Eval}: Multi30k (German)}\\
    \hline    
    
     \rowcolor{LLGray}  \small FT (MT Data)  & \small 0.0 & \small 0.0 & \small 0.0 & \small 0.0  & \small 0.0 & \small 0.0 & \small 0.0   \\
    \small  +Paraphrase  & \small 0.1 & \small 6.8 & \small 20.5 & \small 0.1 & \small 5.3 & \small 22.1 & \small 9.2  \\
    \small +Diverse Recap  & \small 0.2 & \small 7.2 & \small 26.1 & \small 0.5 & \small 7.7 & \small 24.2 & \small 11.0   \\
    \small +Tgt Recap  & \small \textbf{0.8} & \small \textbf{10.5} & \small \textbf{28.8} & \small \textbf{0.8} & \small \textbf{11.1} & \small \textbf{29.8} & \small \textbf{13.6}   \\

     \rowcolor{LLGray} \small FT (Native De) & \small 10.9 & \small 52.2 & \small 100.0   & \small 10.6 & \small 56.2 & \small 100.0    & \small 55.0 \\
    \hline 
    \end{tabular}
    \end{center}
    \vspace{-0.3cm}
    \caption{\textbf{Targeted image recaptioning is especially helpful on error cases which come from not using native speaker text.} Retrieval on STAIR (Ja)/Multi30k (De) \textit{Native vs. Translation Error Sets}.}
    \label{table:nds}
\end{table}

                 \begin{figure*}[t]
    \centering
    \includegraphics[width=\linewidth]{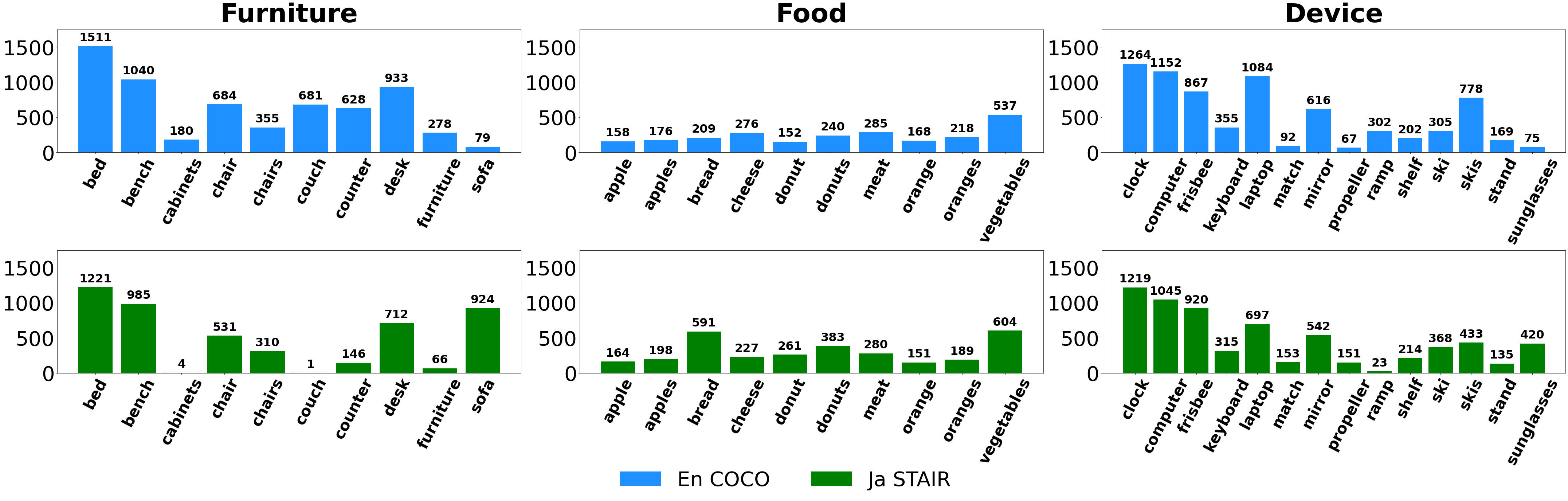}
    \vspace{-7mm}
    \caption{\textbf{When comparing English COCO vs. Japanese STAIR captions, object term distributions are found to vary across languages.} For each supercategory, any term with count $>$ 150 is identified, and the union of terms across languages is shown. Note unique variation across common objects (\emph{e.g.} counter, furniture, bread, sunglasses).}
    
    \label{spec_fig}
\end{figure*} 
         \textbf{Image-text retrieval with native speaker text}. Reported are results on (1) overall STAIR and Multi30k (Tab.~\ref{table:main}) and (2) the \textit{Translation vs. Native Error Sets} which isolate perceptual differences between English and German/Japanese (Tab.~\ref{table:nds}). The best method in both tables 
         is \textit{Targeted Image Recaptioning}. In Tab.~\ref{table:main}, mean recall gains over default finetuning are +2.4 on STAIR and +1.2 on Multi30k. Gains are especially notable for Japanese, which differs much from English. In this case study, the targeted method outperforms \textit{Diverse Paraphrasing} by +1.1 and \textit{Diverse Image Recaptioning} by +0.9, respectively. These results illustrate benefits in using a targeted mechanism with a modest amount of native speaker data ($\approx$10k total). We test other reference set sizes in App. \ref{ref_set}.
        
 \setlength{\tabcolsep}{0.6pt}
\begin{table}
    \begin{center}
    \renewcommand{\arraystretch}{0.82}
    \begin{tabular}{c|ccc|ccc|c}
    \hline
      \textbf{Method}  & \multicolumn{3}{c|}{\textbf{I2T Retrieval}} & \multicolumn{3}{c|}{\textbf{T2I Retrieval}} & \textbf{Mean} \\
      & \small R@1 & \small R@5 & \small R@10  & \small R@1 & \small R@5 & \small R@10 & \textbf{Recall}  \\
    
    \hline
    \multicolumn{8}{c}{\small \textbf{Train:} English COCO / \textbf{Eval}: XM3600 (Japanese)}\\
    \hline
    \rowcolor{LLGray} \small FT (MT Data)  & \small 39.4 & \small 67.6 & \small 78.3 & \small 38.9  & \small 68.5 & \small 78.2 & \small 61.8  \\
    
     \small  +Paraphrase  & \small 41.1 & \small 69.6 & \small 80.0 & \small 41.1 & \small 69.5 & \small 79.5 & \small 63.5   \\
     \small +Diverse Recap & \small 42.6 & \small 70.0 & \small 80.6 & \small 42.1  & \small 70.6 & \small 79.6 & \small 64.2    \\
     \small +Tgt Recap  & \small \textbf{42.9} & \small \textbf{71.6} & \small \textbf{80.8}   & \small \textbf{42.9} & \small \textbf{72.1} & \small \textbf{81.7} & \small \textbf{65.3}  \\

    \hline
    \multicolumn{8}{c}{\small \textbf{Train:} English Flickr30k / \textbf{Eval}: XM3600 (German)}\\
    \hline
     \rowcolor{LLGray} \small FT (MT Data) & \small 38.4 & \small 67.3 & \small 77.4 & \small 37.8  & \small 65.0 & \small 75.9 & \small 60.3  \\
     \small +Paraphrase  & \small 39.1 & \small 68.3 & \small 78.2 & \small 37.3  & \small 64.8 & \small 75.8 & \small 60.6   \\
     \small +Diverse Recap & \small \textbf{39.5} & \small \textbf{68.6} & \small 78.8 & \small 38.0  & \small 65.2 & \small \textbf{77.1} & \small 61.2  \\
     \small +Tgt Recap  & \small \textbf{39.5} & \small 68.5 & \small \textbf{78.9} & \small \textbf{38.7}  & \small \textbf{66.6} & \small 76.6 & \small \textbf{61.5} \\
     

     \hline
    \end{tabular}
    \end{center}
    \vspace{-0.3cm}
    \caption{\textbf{Targeted image captioning is effective cross-dataset.} Shown is retrieval on XM3600 (intra-language) for models trained on Japanese/German.}
    \label{table:xm3600_grouped}
\end{table}

         \noindent \textbf{Error cases which capture perceptual differences}. Tab.~\ref{table:nds} illustrates the value of  \textit{Targeted Image Recaptioning} in addressing perceptual gaps. It outperforms all methods by +4.4 on Japanese and +2.6 on German. The method may perform well on these cases due to enhanced use of culture-specific terms, which we validate with captioning in App. \ref{capt} and term counts in App. \ref{term}. Describing one example, the counts of \textit{bento} for default English captions, paraphrasing, and general captioning are 6, 4, and 5, respectively. The targeted method has 12, closer to the native Japanese 25. We also verify that LLM is not simply hallucinating terms, and show example rewrites in App. \ref{halluc}.

         \noindent \textbf{Across datasets}. To explore the generalizability of the learned perceptual understanding, we test methods cross-dataset on XM3600 image-text retrieval (Tab.~\ref{table:xm3600_grouped}). We find that \textit{Targeted Image Recaptioning} is also the top method cross-dataset. The Japanese model performs especially well, improving by at least +1.1 over all methods. This experiment shows that our method results in understanding that is applicable outside of the training domain.
                   
 \setlength{\tabcolsep}{0.75pt}
\begin{table}
    \begin{center}
    \renewcommand{\arraystretch}{0.82}
    \begin{tabular}{c|c|c|c}
    \hline
      \multicolumn{3}{c|}{\textbf{Rewrites for Image}}  & \textbf{Mean} \\
      \cline{1-3}
        \small Paraphrase & \small Diverse Img Recap & \small Targeted Img Recap  & \textbf{Recall}  \\
    \hline
    &  &  & \small 39.3  \\
    \checkmark &  &  & \small 40.6  \\
    & \checkmark &  & \small 40.8  \\
     &  & \checkmark & \small 41.7  \\
    \checkmark & \checkmark &  & \small 41.6  \\
     & \checkmark & \checkmark & \small 42.5 \\
    \checkmark &  & \checkmark & \small 42.4  \\
     \checkmark & \checkmark & \checkmark & \small \textbf{42.8}  \\
    \hline 
    \end{tabular}
    \end{center}
    \vspace{-0.3cm}
    \caption{\textbf{Targeted image recaptioning is complementary to other augmentation strategies.} Shown is mean recall on STAIR (Ja) when combining rewrite strategies with default translation data (30 epochs).}
    \label{table:combos}
\end{table}

         \noindent \textbf{In combination with other augmentations}. We test each method together by allowing random sampling from combined sets during training (Tab.~\ref{table:combos}). The top gains are from the combination of all methods (+3.5), and next are respective combinations of \textit{Targeted Image Recaptioning} with \textit{Diverse Image Recaptioning} (+3.2) and \textit{Diverse Paraphrasing} (+3.1). In analyzing these complementary benefits, we reason that paraphrasing can address general term diversity shared across speakers of different languages (\emph{e.g.} describing a \textit{car} sometimes as a \textit{vehicle}), and both forms of multimodal recaptioning can help address differences in object focus by incorporating new concepts not in the original caption (\emph{e.g.} adding \textit{clothes} if not mentioned). The targeted method has further advantages by addressing unique properties for a given language (\emph{e.g.} often referring to a \textit{lunch box} as a \textit{bento box}).

        \noindent \textbf{Using other image neighbors.} We test \textit{Targeted Image Recaptioning} using different image neighbors (in terms of similarity) to guide rewrites. On STAIR, using $k$=1 results in 41.7 mean recall, $k$=2 results in 41.6, and $k$=3 results in 41.6. These results suggest that other neighbors can be effective and add further diversity.

        \begin{figure}[t]
    \centering
    \includegraphics[width=1\linewidth]{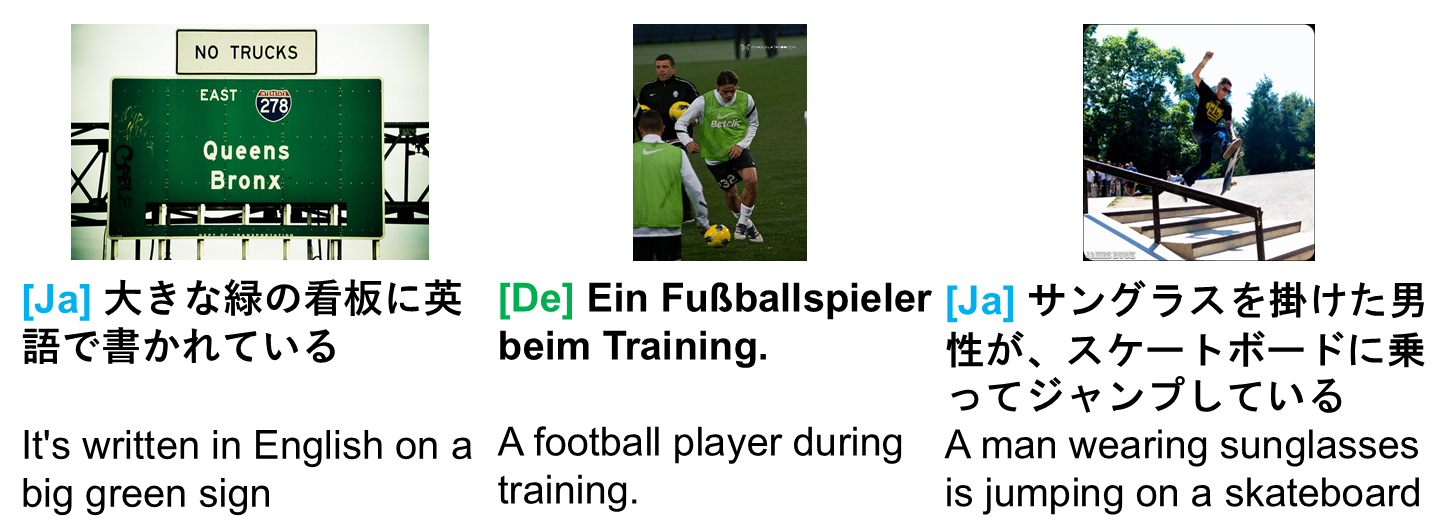}
    \vspace{-7mm}
    \caption{\textbf{I2T retrievals that \textit{Targeted Image Recaptioning} gets correct but default finetuning gets incorrect@10.} The targeted method addresses unique differences in perspective and level of detail.
    }
    \label{qual}
\end{figure}

        \subsection{How do differences in object descriptions uniquely manifest across languages?} 

        There are culture-specific term frequency differences expected across languages (\textit{e.g.} \textit{futon} appearing more in Japanese). However, less obvious are differences in the distributions of common nouns. To quantify such differences, we use the method in Sec.~\ref{recipe}, and calculate supercategory-grouped counts of common objects for Japanese and German vs. English. In Fig. \ref{spec_fig}, we show examples for \textit{furniture}, \textit{food}, and \textit{device} (Japanese vs. English). Results for German vs. English and for other categories are in Appendix \ref{app_dist}. 
        
        In Fig. \ref{spec_fig}, for the supercategory \textit{furniture}, the En set uses the term \textit{furniture} 4.2$\times$ the Ja set while Ja describes \textit{sofa}/\textit{couch} 1.2$\times$, showing potential differences in object grouping. For \textit{food}, \textit{bread} is described 2.8$\times$ more in Ja. Upon inspection, we find phrases like \textit{bread with meat} are used synonymously with \textit{sandwich}. For \textit{device}, \textit{sunglasses} is described 5.6$\times$ more in Ja, potentially a result of sunglasses culturally being less common in Japan (and more noteworthy). Cases like these point to perceptual diversity that is worthy of future study. 

        Further examples show up in the retrieval error cases. Shown in Fig. \ref{qual} are some I2T retrieval cases in the \textit{Translation vs. Native Error Sets} for Japanese/German that \textit{Targeted Image Recaptioning} gets correct, but the finetuned baseline does not. Observe how the targeted method improves on unique cases, such as an out-group view of a New York sign, and the text describing \textit{sunglasses} (a description difference identified in Fig. \ref{spec_fig}).

 \setlength{\tabcolsep}{1.7pt}
\begin{table}
    \begin{center}
    \renewcommand{\arraystretch}{0.85}
    \begin{tabular}{c|ccccccc}
    \hline
      \small \textbf{Method}  & \small \textbf{Ja} & \small \textbf{Ko} & \small \textbf{Zh} & \small \textbf{De} & \small \textbf{Fr} & \small \textbf{Cs} & \small \textbf{Da} \\

    \hline
    \multicolumn{8}{c}{\small \textbf{Train:} English COCO to Japanese / \textbf{Eval}: XM3600 }\\
    \hline

    \rowcolor{LLGray} \small FT (MT Data)  & \small \cellcolor{LLGray}61.8 & \small 23.6 & \small 51.9 & \small 55.8 & \small 34.7 & \small 42.7 & \small 34.0 \\
    
     \small  +Paraphrase & \small \cellcolor{LLGray}63.5& \small 25.9 & \small 54.5 & \small 57.1 & \small 36.3 & \small 44.8 & \small 36.5 \\
     \small +Diverse Recap & \small \cellcolor{LLGray}64.2 & \small 24.8 & \small \textbf{54.8} & \small 58.2 & \small 36.6 & \small \textbf{45.6} & \small 36.3  \\
     \small +Tgt Recap & \small \cellcolor{LLGray}\textbf{65.3} & \small \textbf{26.0} & \small \textbf{54.8} & \small \textbf{58.3} & \small \textbf{37.3} & \small 44.8 & \small \textbf{36.8} \\

    \hline
    \multicolumn{8}{c}{\small \textbf{Train:} English Flickr30k to German / \textbf{Eval}: XM3600}\\
    \hline
      \rowcolor{LLGray} \small FT (MT Data)  & \small 54.6 & \small 21.6 & \small 49.8 & \small \cellcolor{LLGray}60.3 & \small 34.8 & \small 42.6 & \small 37.3  \\
    
     \small  +Paraphrase & \small \textbf{55.5} & \small 21.7 & \small \textbf{50.7} & \small \cellcolor{LLGray}60.6 & \small \textbf{36.1} & \small 43.9 & \small 38.3 \\
     \small +Diverse Recap& \small 55.1 & \small \textbf{22.8}  & \small 50.3 & \small \cellcolor{LLGray}61.2 & \small 36.0 & \small \textbf{44.1} & \small \textbf{38.6}  \\
     \small +Tgt Recap & \small 54.7 & \small 22.2 & \small 50.6 & \small \cellcolor{LLGray}\textbf{61.5} & \small 35.5 & \small 43.9 & \small 38.3  \\

     \hline
    \end{tabular}
    \end{center}
    \vspace{-0.3cm}
    \caption{\textbf{There is a need to learn about perceptual diversity in other languages.} We report mean recall for retrieval on different language-specific sets when performing targeted recaptioning for Japanese/German. These results show targeted recaptioning to be best intra-language, but other strategies to be similarly (or more) compelling cross-language. Perceptual diversity learned from Japanese/German may thus be unique.}
    \label{table:xm3600_cross_lang}
\end{table}

        \subsection{Does perceptual diversity understanding gained from one target language dataset generalize to other datasets/languages?} 

        In addition to cross-dataset tests with XM3600, we also test models in a cross-lingual manner by performing targeted recaptioning for one language (Japanese/German) and evaluating on language-specific retrieval sets for geographically proximate languages: Korean/Chinese for Japanese and French/Czech/Danish for German. Note that high performance in each case is \textit{not} a goal of our model, but this study provides insight into whether some perceptual understanding may be generalizable across languages. Table ~\ref{table:xm3600_cross_lang} shows the results. While the targeted method is best intra-language, the other rewrite strategies  become more compelling cross-language, with smaller gaps or gains vs. the targeted method. These results indicate that other languages may not benefit much from learning German/Japanese perceptual details, and likely have their own unique perceptual diversity. This insight can inspire future work to ensure adequate consideration of native speaker data from other languages.

\section{Conclusion}

    In this work, we provide a multimodal framework to encourage VLMs to learn diverse perceptual understanding across languages. We find that multimodal LLMs, with nearest-neighbor guidance, are effective at inferring how to change object descriptions across languages. Our targeted method improves image-text retrieval, especially on error cases that come from English perceptual bias. The targeted method is also found to be complementary to other augmentations, and gains generalize across datasets. We provide unique insights into obvious and more subtle ways that object text differences manifest across cross-language datasets. Future work needs to be dedicated to the acquisition of native speaker captions across other languages for more expansive investigation.  

    \noindent \textbf{Acknowledgement.} This work was supported by NSF Grant 2329992 and a University of Pittsburgh Intelligent Systems Provost Fellowship.

\section{Limitations}

    First, our framework is limited by the availability of native speaker image captions across languages. This constraint is the primary reason we study Multi30k German and STAIR Japanese, since they have captions directly produced from native speakers to pair with English text of Flickr and COCO, respectively. We encourage the acquisition of native speaker data from more languages, especially low-resource ones, for study in the future.

    Second, we use a single, small set of reference examples. While the mechanism is performant and data-efficient, there is intra-language diversity that remains uncaptured in such a set. Furthermore, the domain shift between the reference set and a test set of interest can be significant. Future work can expand the language and image diversity represented in the reference set to address such cases. 

    Third, our framework and analysis depend on machine translation quality. While we verify effectiveness across multiple machine translation techniques, improvements in machine translation are likely needed to maximize cross-lingual performance. Some object description count differences may be a result of translation artifacts (\emph{e.g.} \textit{sofa}/\textit{couch}), though we combine cases in analysis. 
    
    Fourth, our WordNet mechanism to identify object description differences across datasets is imperfect due to the idiosyncratic synset structure of WordNet. Polysemy can affect interpretation of counts, through we try to handle some aspects of disambiguation (\emph{e.g.} differentiating between noun and adjective forms of \textit{orange} through part-of-speech tagging). Future methods can be developed to probe differences more precisely.  
    
    Fifth, we only explore one model each for recaptioning and retrieval training, but models of different scale may show different behaviors. We reason larger models with stronger instruction following may be more effective at altering input captions to reflect the guidance examples. They may be able to identify the most relevant differences between languages to use for adaptation, and even discover subtle differences. In addition, there may be less risk of hallucination. Conversely, smaller models with less task-following capability may be less effective at incorporating desired changes. Future work can investigate the impact of scaling on a model's ability to understand perceptual diversity. Retrieval models like SigLIP \citep{zhai2023sigmoid} may be worth studying.

\section{Ethical Considerations}

    While we address one form of bias, there are various other biases in the datasets we use for training and evaluation, such as racial and gender biases. An extra filtering or rewriting step could potentially help address these biases for downstream use cases. Our targeted mechanism could  also help generate training data to teach models less biased descriptions of general content. 

\bibliography{custom}

\appendix

\section{Appendix}

    \label{sec:appendix}

    \subsection{Prompts for Rewrite Strategies}
        \label{app_prompt}

    \begin{quote}
        \textbf{Diverse Paraphrasing} \\
        Task: The objective is to paraphrase an English caption to reflect diversity in how speakers around the world describe objects, especially across languages. It is very important to strictly follow the listed requirements. 

        Requirements: \\
        - Output only a single paraphrased caption which must start with <final> and end with </final>.\\
        - Example: <final> There is a blue bicycle and red motorcycle on the street. </final>\\
        - Do not output any additional quotes, text, comments, explanations, or details. Just the caption.
        
        Please complete this example: \\
        Input: \{input\}\\
        Output: 
    \end{quote}

    \begin{quote}
        \textbf{Diverse Image Recaptioning} \\
        Task Description: For an input image and an input caption, produce a one-sentence image caption that differs significantly from the input caption in order of phrases, sentence structure, semantic content, which objects are described, and/or level of detail. Make sure the output differs from the input caption and use the image for guidance. Only perform changes that are correct and semantically relevant to the given input image. After "Output: ", always output a <final> tag, followed by a rewritten caption, then </final>. Never any other text or explanation. One task demo for formatting and change instruction is provided. 

        Task Demo:        
        Inference \\
        Input: A young boy holding a baseball bat during a baseball game. \\
        Output: <final> The batter in the grey uniform is waiting for a ball during a game. </final>
        
        Now perform the task exactly as above:        
        Inference\\
        Input: \{input\}\\ 
        Output: 
    \end{quote}

    \begin{quote}
        \textbf{Targeted Image Recaptioning} \\
        Task Description: For an input image, image caption, and reference input-output caption(s) for similar image(s), rewrite the image caption with similar changes to the style, level of detail, and object terms as in the reference examples. Only perform changes that are correct and semantically relevant to the given input image. After "Output: ", always output a <final> tag, followed by a rewritten caption, then </final>. Never any other text or explanation. One task demo for formatting and change instruction is provided. 

        Task Demo:\\
        Reference example(s)\\
        Input: A catcher catching a ball that has just gone by the hitter.\\
        Output: The batter in the orange uniform just missed the ball. \\
        Inference\\
        Input: A young boy holding a baseball bat during a baseball game.\\
        Output: <final> The batter in the grey uniform is waiting for a ball during a game. </final>
        
        Now perform the task exactly as above:        
        Reference example(s)\\
        \{reference\_examples\}
        
        Inference \\
        Input: \{input\} \\
        Output: 

    \end{quote}

    \subsection{Evaluation of NLLB Translation Model}
        \label{nllb}

        We consider another translation model, No Language Left Behind \citep{costa2022no}. Results after training with translation to Japanese are shown in Tab. \ref{table:nllb}. The results trail those of the HelsinkiNLP model, potentially a result of the HelsinkiNLP model being language-dedicated, while No Language Left Behind is focused on multilinguality. These results nonetheless show that the targeted method works well across translators.

 \setlength{\tabcolsep}{0.6pt}
\begin{table}[t]
    \begin{center}
    \renewcommand{\arraystretch}{0.83}
    \begin{tabular}{c|ccc|ccc|c}
    \hline
      \textbf{Method}  & \multicolumn{3}{c|}{\textbf{I2T Retrieval}} & \multicolumn{3}{c|}{\textbf{T2I Retrieval}} & \textbf{Mean} \\
      & \small R@1 & \small R@5 & \small R@10  & \small R@1 & \small R@5 & \small R@10 & \textbf{Recall}  \\
    
    \hline
    \multicolumn{8}{c}{\small \textbf{Translation:} HelsinkiNLP Tatoeba (to Japanese)}\\
    \hline
    
\rowcolor{LLGray} \small FT (MT Data)  & \small 19.4 & \small 41.9 & \small 53.3 &  \small 20.2 & \small 43.0  & \small 54.7  & \small 39.3   \\
    
     \small  +Paraphrase &  \small 21.7 & \small 45.1  & \small 56.7 & \small 20.6 & \small 43.9 & \small 55.4 & \small 40.6   \\
     \small +Diverse Recap  &  \small 22.0 & \small 45.3  & \small 57.1& \small 20.7 & \small 44.1 & \small 55.6  & \small 40.8  \\  
     \small \textbf{+Tgt Recap} &  \small \textbf{22.5} & \small \textbf{46.5}  & \small \textbf{58.1}  & \small \textbf{21.4} & \small \textbf{45.0} & \small \textbf{56.7}  & \small \textbf{41.7}  
    \\
    \hline
    \multicolumn{8}{c}{\small \textbf{Translation:} No Language Left Behind (to Japanese)}\\
    \hline
    
    \rowcolor{LLGray} \small FT (MT Data)  & \small 19.5 & \small 41.7 & \small 52.8 & \small 18.6  & \small 40.4 & \small 51.9 & \small 37.5   \\
    
     \small  +Paraphrase & \small 20.7 & \small 43.6 & \small 54.6 & \small 19.7  & \small 42.1 & \small 53.4 & \small 39.0   \\
     \small +Diverse Recap  & \small 21.2 & \small 44.2 & \small 55.5 & \small 20.0  & \small 42.6 & \small 54.0 & \small 39.6   \\
     \small \textbf{+Tgt Recap}  & \small \textbf{21.6} & \small \textbf{45.0} & \small \textbf{56.5} & \small \textbf{20.5}  & \small \textbf{43.6} & \small \textbf{55.2} & \small \textbf{40.4} \\
    \hline 
    \end{tabular}
    \end{center}
    \vspace{-0.3cm}
    \caption{\textbf{The targeted method achieves gains across translation models, and HelsinkiNLP Tatoeba outperforms NLLB}. Translation is performed on English COCO, and evaluation is on Japanese STAIR.}
    \label{table:nllb}
\end{table}

         \setlength{\tabcolsep}{0.6pt}
\begin{table}[t]
    \begin{center}
    \renewcommand{\arraystretch}{0.83}
    \begin{tabular}{c|c|c}
    \hline
      \textbf{Method} & \textbf{Reference Set Size} & \textbf{Mean Recall} \\
    
    \hline
    \rowcolor{LLGray} \small FT (MT Data)  & \small - & \small 39.3    \\
    
     \small +Tgt Recap  &  \small 4,833 & \small 41.5   \\
     \small +Tgt Recap & \small 9,666 & \small 41.7  \\  
     \small +Tgt Recap & \small 19,332 & \small 41.8  \\ 
    \hline 
    \end{tabular}
    \end{center}
    \vspace{-0.3cm}
    \caption{\textbf{The targeted method results in retrieval gains across reference set sizes}. Recaptioning here uses varying \# of English COCO images (and corresponding STAIR captions) in the reference set. We report results with a reference set of 9,666 examples throughout the paper, but 4,833 examples is shown to also be effective. Evaluation is on Japanese STAIR.}
    \label{table:ref_set}
\end{table}
         \setlength{\tabcolsep}{0.6pt}
\begin{table*}[t]
    \begin{center}
    \renewcommand{\arraystretch}{0.83}
    \begin{tabular}{c|c|c|c|c|c}
    \hline
      \small \textbf{Method} & \small \textbf{ROUGE-1} & \small \textbf{ROUGE-2} & \small \textbf{ROUGE-3} & \small \textbf{ROUGE-4} & \small \textbf{ROUGE-L} \\
    
    \hline
    \rowcolor{LLGray} \small En Caps &  \small 0.332 & \small 0.095 & \small 0.031 & \small 0.011 & \small 0.293    \\
    
     \small +Paraphrase  &  \small 0.321 & \small 0.085 & \small 0.029 & \small 0.010 & \small 0.278   \\
     \small +Diverse Recap &  \small 0.363 & \small 0.110 & \small 0.042 & \small 0.018 & \small 0.295   \\  
     \small \textbf{+Tgt Recap}&  \small \textbf{0.369} & \small \textbf{0.128} & \small \textbf{0.052} & \small \textbf{0.024} & \small \textbf{0.323}  \\ 
    \hline 
    \end{tabular}
    \end{center}
    \vspace{-0.3cm}
    \caption{\textbf{Targeted image recaptioning is the most effective strategy in terms of captioning metrics, ROUGE-1/2/3/4/L (avg. F1 scores)}. The reference set here is Japanese STAIR train (translated to English). The rewrites, before translation to Japanese, are evaluated versus the references.}
    \label{table:cap}
\end{table*}

    \subsection{Impact of Reference Set Size}
        \label{ref_set}

         The reference set size is a key parameter that could affect the quality of rewrites and thus retrieval performance. Our Japanese STAIR experiments in Table \ref{table:main} demonstrate that a training set of size \textasciitilde73k images can benefit from a smaller reference set of size \textasciitilde10k images. To provide sensitivity analysis, we additionally test a smaller \textasciitilde5k reference set and a larger \textasciitilde20k reference set for the same training set size (\textasciitilde73k), sampling extra reference images and captions from previously unused COCO examples. The results are shown in Table \ref{table:ref_set}. We find a smaller reference set (5k) to be effective vs. the baseline (41.5 vs. 39.3), indicating further opportunity for data efficiency with our method. We find a larger reference set (20k) to be effective vs. the baseline (41.8 vs. 39.3), though perform just slightly better than the 10k set (41.8 vs. 41.7), implying diminishing returns. 

    \subsection{Captioning Evaluation}
        \label{capt}

        While our focus is on text-image retrieval, we provide an alternative evaluation of targeted image recaptioning in terms of captioning metrics. We particularly provide a small-scale captioning comparison of the original English training captions and the output captions from each rewrite strategy versus a set of references consisting of captions from the Japanese STAIR training set (which is unseen in the recaptioning process). This comparison is designed to gauge if the targeted rewrites more closely align the native Japanese captions vs. the other baselines in terms of the words and descriptions used. 
        
        Of note, the rewrites are first output from the multimodal LLM in English (see Fig. \ref{prompt_fig}). With focus on measuring syntactic overlap, which can be done in English, we simply translate Japanese references to English (Google Translate). Then we score in terms of ROUGE-1/2/3/4 and ROUGE-L (avg. F1 scores).  Results are shown in Table \ref{table:cap}. We find our targeted recaptioning to be the most effective strategy across all metrics, with results indicating that the targeted rewrites most closely align the syntactic structure of the native Japanese captions. These results can inspire more directed captioning work in the future. 

    \subsection{Term Count Comparisons}
        \label{term}

 \setlength{\tabcolsep}{1.7pt}
\begin{table*}
    \begin{center}
    \renewcommand{\arraystretch}{0.8}
    \begin{tabular}{c|cccccccc}
    \hline
      \small \textbf{Method}  & \small \textbf{platform} & \small \textbf{bento} & \small \textbf{futon} & \small \textbf{sunglasses} & \small \textbf{car} & \small \textbf{western} & \small \textbf{jumper} 
        & \small \textbf{ramen} \\
    \hline

    \rowcolor{LLGray} \small Native En  & \small 147  & \small 6 & \small 5 & \small 77 & \small 847 & \small 3 & \small 1 & \small 0 \\
     \small  Paraphrase & \small 251  & \small 4 & \small 0 & \small 74 & \small 370 & \small 3 & \small 2 & \small 1 \\
     \small Diverse Recap & \small 162  & \small 5 & \small 5 & \small 62 & \small 818 & \small 1 & \small 0 & \small 3 \\
     \small Tgt Recap & \small 309  & \small 12 & \small 15 & \small 145 & \small 1077 & \small 52 & \small 6 & \small 6 \\
     \rowcolor{LLGray} \small Native Ja & \small 491  & \small 25 & \small 87 & \small 422 & \small 1379 & \small 181 & \small 36 & \small 9 \\

     \hline
    \end{tabular}
    \end{center}
    \vspace{-0.3cm}
    \caption{\textbf{With the targeted method, term counts get closer to the distribution of native Japanese.} Shown are counts in the training sets for each of the methods. Notice that the terms cover culture-specific naming, unique salient content, and different perspectives. }
    \label{table:terms_ja}
\end{table*}

 \setlength{\tabcolsep}{1.7pt}
\begin{table*}
    \begin{center}
    \renewcommand{\arraystretch}{0.8}
    \begin{tabular}{c|cccccccc}
    \hline
      \small \textbf{Method}  & \small \textbf{rugby} & \small \textbf{motorcyclist} & \small \textbf{lectern} & \small \textbf{football} & \small \textbf{zone} & \small \textbf{blonde} & \small \textbf{shepherd} & \small \textbf{formula} \\

    \hline

    \rowcolor{LLGray} \small Native En  & \small 7  & \small 5 & \small 1 & \small 63 & \small 1 & \small 0 & \small 10 & \small 0  \\
     \small  Paraphrase & \small 6  & \small 9 & \small 0 & \small 84 & \small 2 & \small 16 & \small 9 & \small 0  \\
     \small Diverse Recap & \small 6  & \small 8 & \small 0 & \small 53 & \small 1  & \small 20 & \small 5 & \small 0 \\
     \small Tgt Recap & \small 13  & \small 20 & \small 3 & \small 82 & \small 10 & \small 33 & \small 11 & \small 3  \\
     \rowcolor{LLGray} \small Native De & \small 19  & \small 20 & \small 4 & \small 94 & \small 14 & \small 52 & \small 15  & \small 2 \\

     \hline
    \end{tabular}
    \end{center}
    \vspace{-0.3cm}
    \caption{\textbf{With the targeted method, term counts get closer to the distribution of native German.} Shown are counts in the training sets for each of the methods. Notice that the terms cover culture-specific naming, object grouping, and unique salient content.}
    \label{table:terms_de}
\end{table*}

        \begin{figure*}[t]
    \centering
    \includegraphics[width=0.96\linewidth]{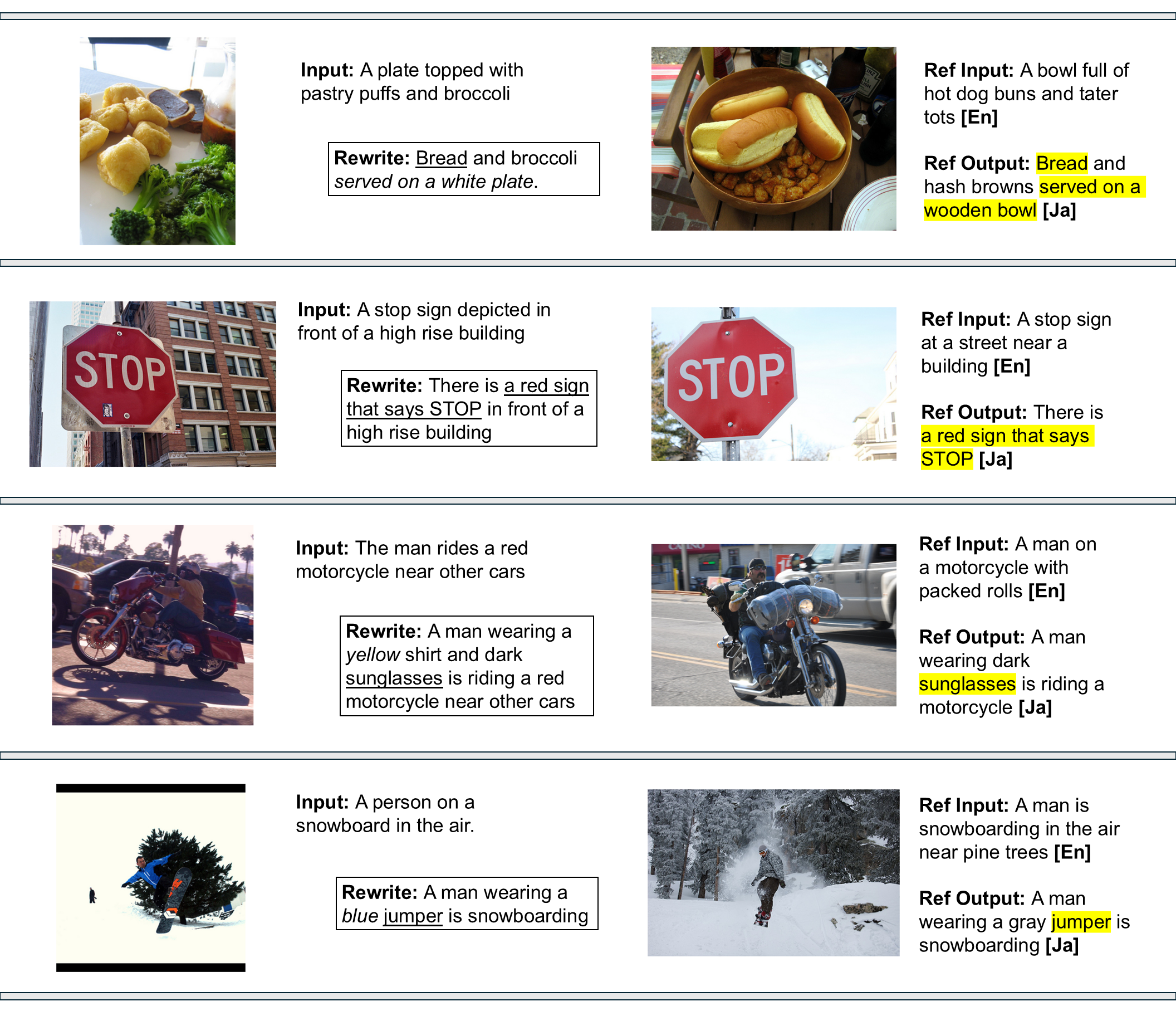}
    \vspace{-3mm}
    \caption{\textbf{Example inputs, nearest-neighbor reference images/captions, and rewrites produced with our targeted image recaptioning (Japanese).} Observe how the model can leverage text in the outputs of the reference (\emph{e.g.} \textit{bread}, \textit{red sign that says STOP}, \textit{sunglasses}, and \textit{jumper}), while inferring relevant details for the input images (\emph{e.g.} the fact that the \textit{jumper} is \textit{blue}). The language in brackets is the language in which the caption was produced. 
    }
    \label{ex_rew}
\end{figure*}

        \begin{figure*}[t]
    \centering
    \includegraphics[width=0.96\linewidth]{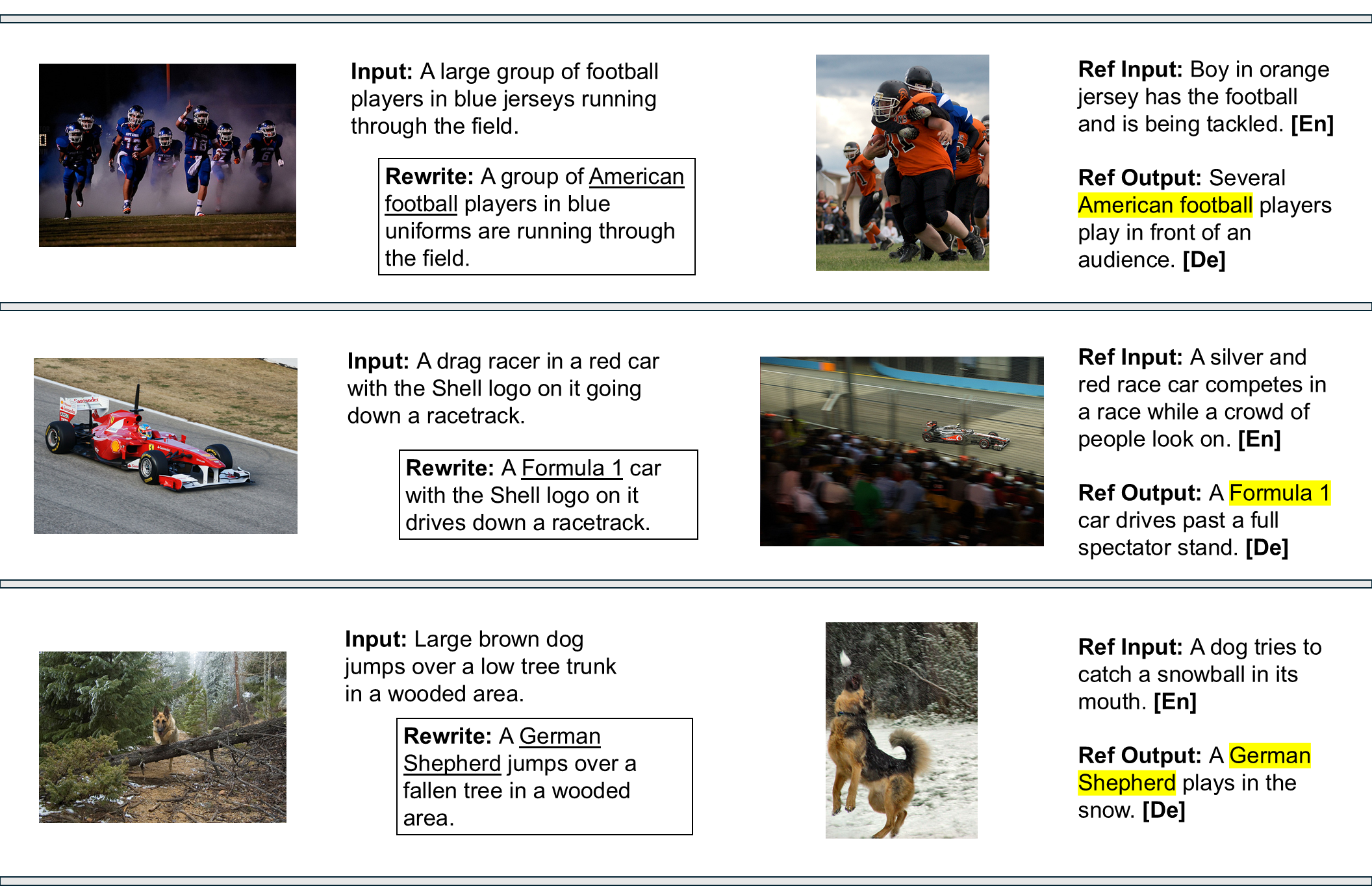}
    \vspace{-3mm}
    \caption{\textbf{Example inputs, nearest-neighbor reference images/captions, and rewrites produced with our targeted image recaptioning (German).} Observe how the model can leverage text in the outputs of the reference (\emph{e.g.} \textit{American football}, \textit{Formula 1}, and \textit{German Shepherd}), while inferring relevant details for the input images (\emph{e.g.} jumping over a \textit{fallen tree}). The language in brackets is the language in which the caption was produced. 
    }
    \label{ex_rew_de}
\end{figure*}

        
        To validate that the targeted recaptioning is capturing object description properties of a target language, we compare term counts of training rewrites to native English and the native target language. Tab. \ref{table:terms_ja} shows examples for Japanese, and Tab. \ref{table:terms_de} shows examples for German. The targeted method successfully accounts for terms which the other augmentation strategies and original English captions do not account for as well. For why these differences exist, future study is needed. We hypothesize that certain terms may be more salient and/or noteworthy. For Japanese,  \textit{sunglasses} are uncommon and perhaps noteworthy. For German, \textit{formula} encompasses ``Formula 1'' racing, which is popular in Europe. Additionally, \textit{zone} encompasses ``pedestrian zones'', which are popular in Germany.

    \subsection{Consideration of Hallucination}
        \label{halluc}

        One potential concern about recaptioning is hallucination of concepts from the LLM.  To mitigate potential hallucination, we encourage the LLM to reason about correct changes, where our prompt states, “Only perform changes that are correct and semantically relevant to the given input image”. To conduct quality evaluation, we perform analysis for targeted image recaptioning. We examine a random sample of 200 generated captions, and find that 94.5\% of the altered captions are entirely correct given the image. These results demonstrate that our method has limited negative impact from hallucination. Instead, the targeted process generally respects the details of the new image, while generalizing how objects should be described using reference captions from native speakers. Some examples are shown in Fig. \ref{ex_rew} and Fig \ref{ex_rew_de}.

        In caption generations with some degree of hallucination/incorrectness, the errors typically involve the addition of one object which does not exist in the given image. This happens when the object to be captioned is really small (\textit{e.g.} “bird” is hallucinated when there is a “person” in the distance), or if there is fine-grained understanding required (\textit{e.g.} understanding the difference between “third base stands” and “first base stands” at a baseball game). We expect that future improvements to multimodal LLMs will overcome these issues.

    \subsection{More Object Description Distributions}
        \label{app_dist}

\begin{figure*}[t]
    \centering
    \includegraphics[width=\linewidth]{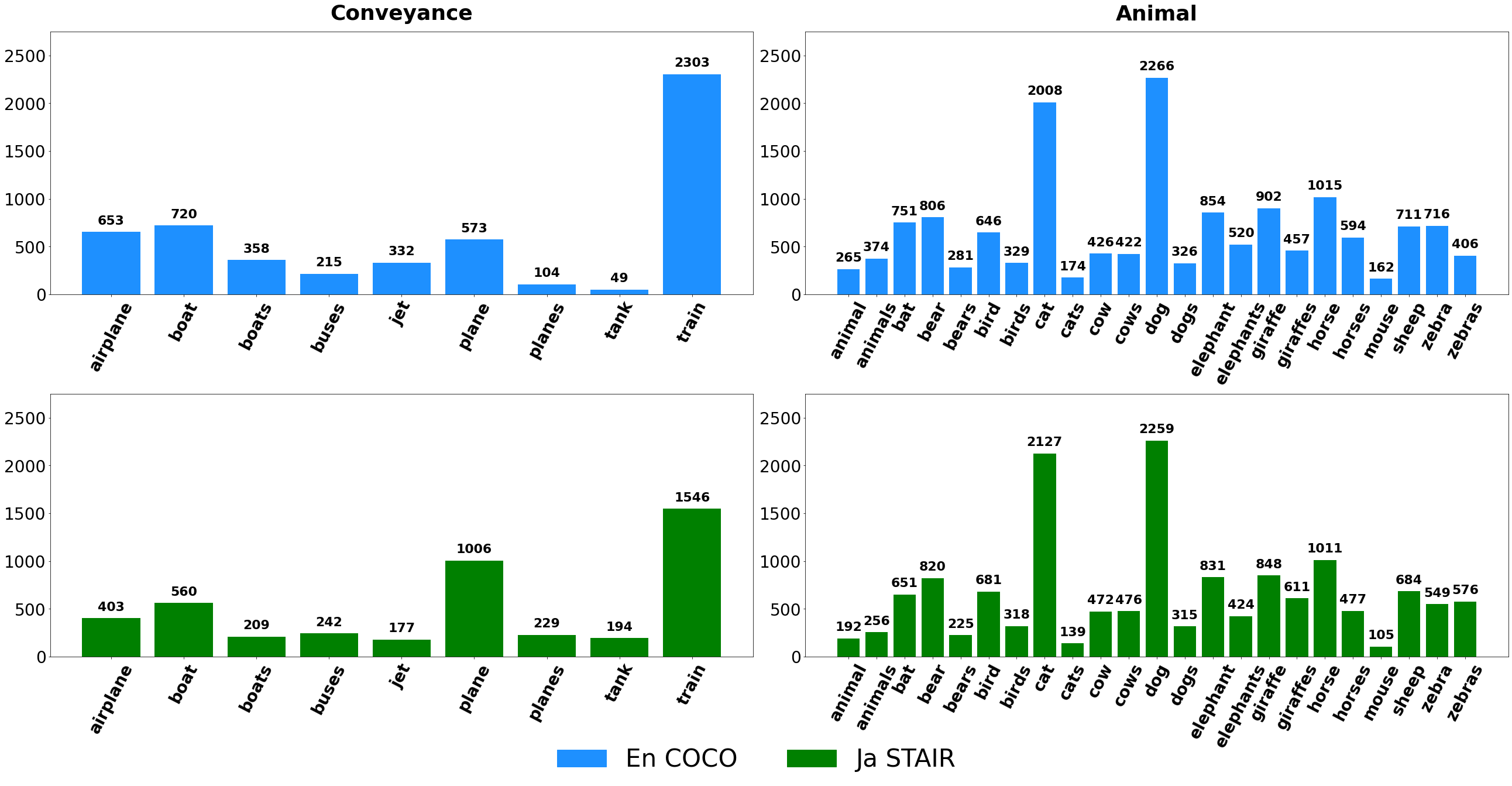}
    \caption{\textbf{Term distributions for \textit{conveyance} and \textit{animal}, English COCO vs. Japanese STAIR.} For each supercategory, any term with count $>$ 150 is identified, and the union of terms across languages is shown.  }
    \label{coco_ext1}
\end{figure*}

\begin{figure*}[t]
    \centering
    \includegraphics[width=\linewidth]{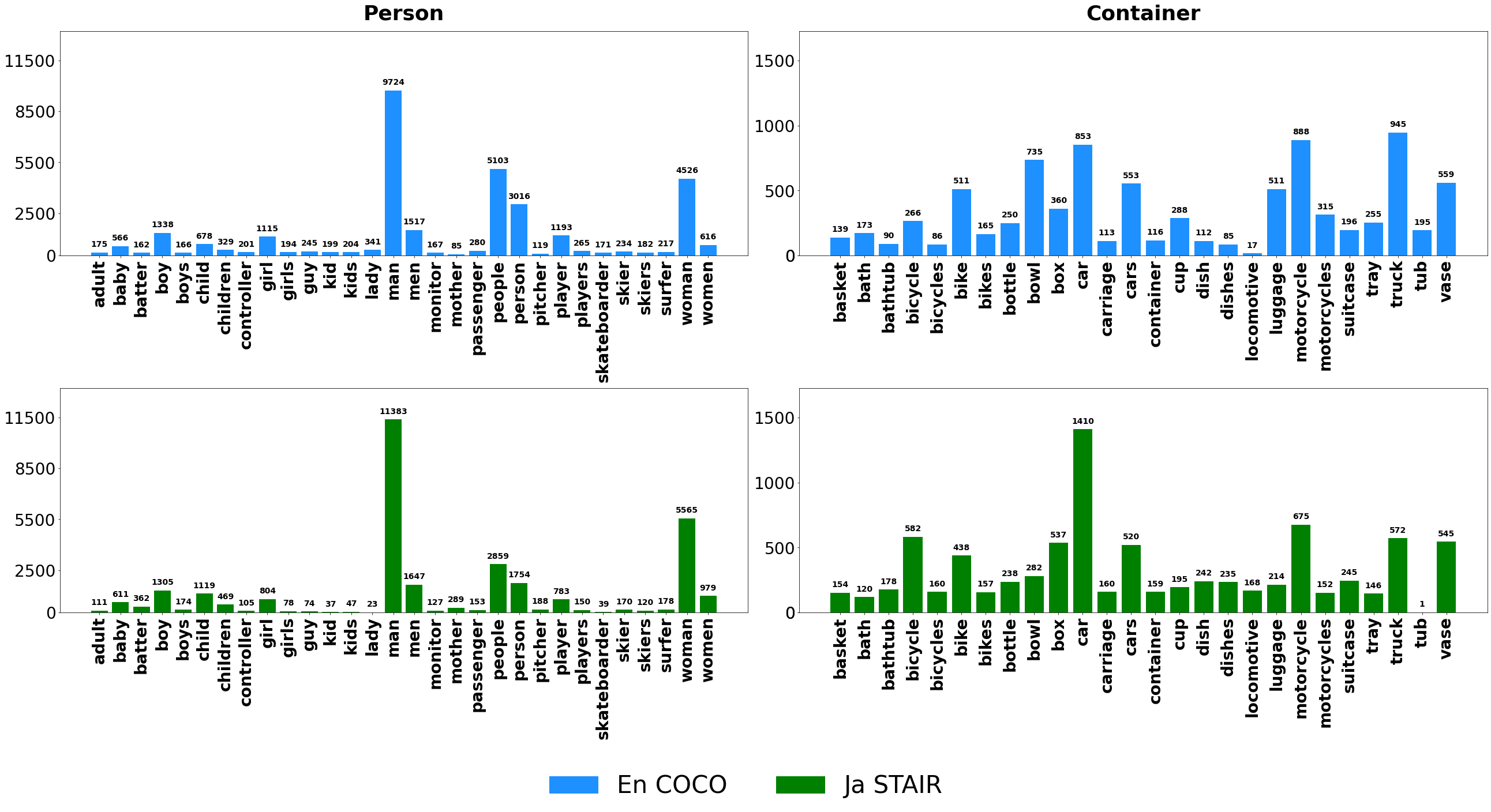}
    \caption{\textbf{Term distributions for \textit{person} and \textit{container}, English COCO vs. Japanese STAIR.} For each supercategory, any term with count $>$ 150 is identified, and the union of terms across languages is shown.}
    \label{coco_ext2}
\end{figure*}

        With respect to Japanese vs. English, we produce more distributions like Fig. \ref{spec_fig} (using one of our Japanese STAIR and English COCO train sets). Fig. \ref{coco_ext1} shows \textit{conveyance} and \textit{animal}, while Fig. \ref{coco_ext2} shows results for \textit{person} and \textit{container}. There are some similarities in counts (\emph{e.g.} the distributions for \textit{animal} are similar in Figure \ref{coco_ext1}). There are also notable differences. In Fig. \ref{coco_ext1}, \textit{jet} is used more often in the En set, while \textit{plane} is used more frequently in the Ja set. In Fig. \ref{coco_ext2}, \textit{locomotive} and \textit{car} are used much more frequently in the Ja set, while \textit{truck} is used more in the En set. Also in Fig. \ref{coco_ext2}, \textit{man} and \textit{woman} are described more in the Ja set, while the En set uses \textit{people} more. 
        
\begin{figure*}[t]
    \centering
    \includegraphics[width=\linewidth]{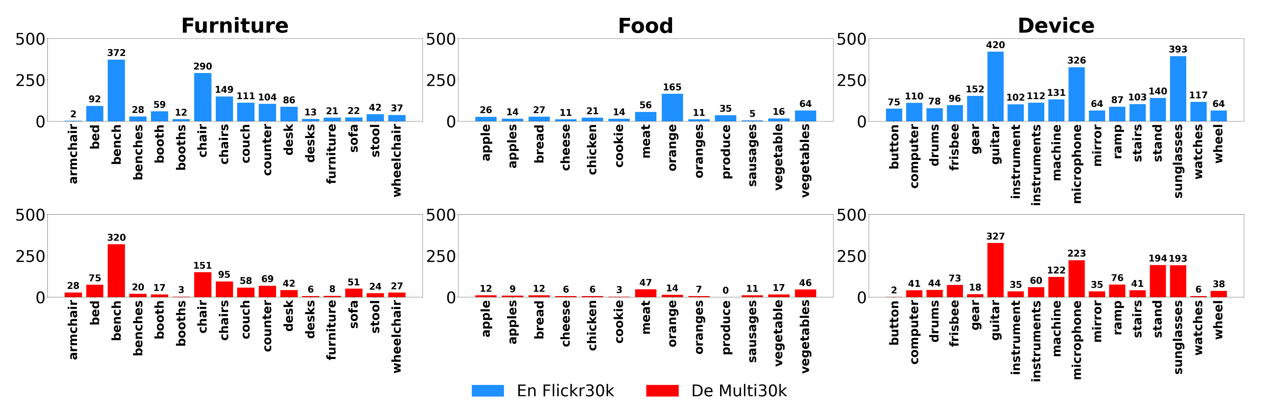}
    \caption{\textbf{Term distributions for \textit{furniture}, \textit{food}, and \textit{animal}, English Flickr30k vs. German Multi30k.} For \textit{furniture}, any term with count $>$ 10 is identified. For \textit{food}, any term with count $>$ 10 is identified. For \textit{device}, any term with count $>$ 60 is identified. Then the union of terms across languages is shown.}
    \label{multi30k_ext1}
\end{figure*}

\begin{figure*}[t]
    \centering
    \includegraphics[width=\linewidth]{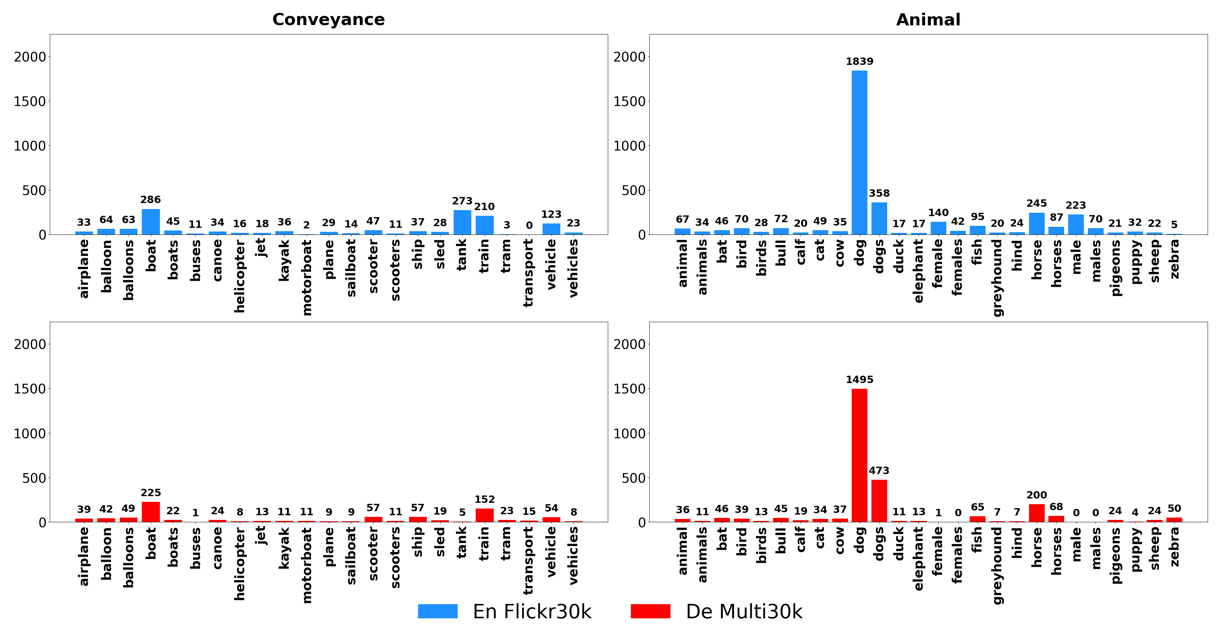}
    \caption{\textbf{Term distributions for \textit{conveyance} and \textit{animal}, English Flickr30k vs. German Multi30k.} For \textit{conveyance}, any term with count $>$ 10 is identified. For \textit{animal}, any term with count $>$ 15 is identified. Then the union of terms across languages is shown.}
    \label{multi30k_ext2}
\end{figure*}

\begin{figure*}[t]
    \centering
    \includegraphics[width=\linewidth]{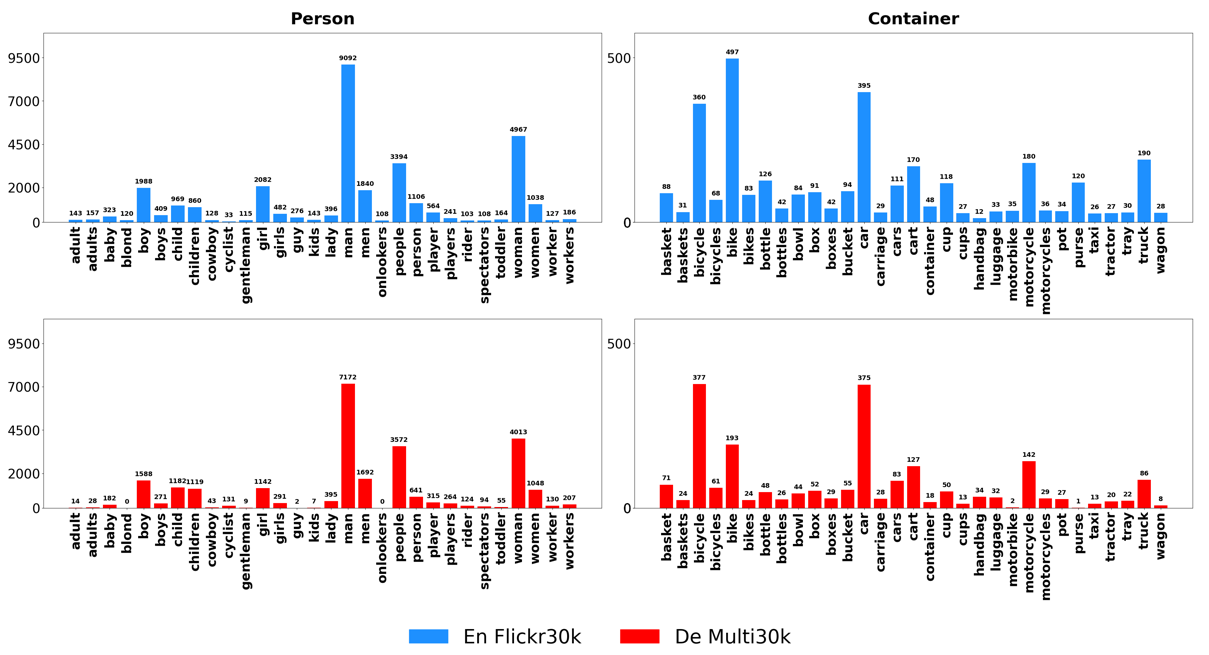}
    \caption{\textbf{Term distributions for \textit{person} and \textit{container}, English Flickr30k vs. German Multi30k.} For \textit{person}, any term with count $>$ 100 is identified. For \textit{container}, any term with count $>$ 25 is identified. Then the union of terms across languages is shown.}
    \label{multi30k_ext3}
\end{figure*}

        Fig. \ref{multi30k_ext1}-\ref{multi30k_ext3} also show comparisons for German vs. English, using one of the original speaker train sets from each Multi30k and Flickr30k. In general, there is more similarity between English and German, which is intuitive considering the proximity of the languages. However, some uniqueness exists. For example for the supercategory \textit{person}, \textit{cyclist} occurs \textasciitilde4$\times$ more often in German than English, through \textit{cowboy} occurs \textasciitilde3$\times$ more often in English. For \textit{device}, \textit{stand} is \textasciitilde1.5$\times$ more frequent in German. Then for the supercategory \textit{container}, \textit{bike} is mentioned \textasciitilde2.5$\times$ less often in German.



\end{document}